\documentclass[fleqn,10pt]{wlpeerj}

\makeatletter
\g@addto@macro\normalsize{%
  \setlength\abovedisplayskip{4pt plus 4pt minus 3pt}%
  \setlength\belowdisplayskip{4pt plus 4pt minus 3pt}%
  \setlength\abovedisplayshortskip{4pt plus 3pt minus 2pt}%
  \setlength\belowdisplayshortskip{4pt plus 3pt minus 2pt}%
}
\makeatother

\usepackage{threeparttable}
\usepackage{array}
\usepackage{xcolor}
\usepackage{comment}
\usepackage{amsmath,amssymb,amsfonts}  
\usepackage{algorithm}
\usepackage{algpseudocode}
\usepackage{float}
\usepackage{booktabs}
\usepackage{multirow}
\usepackage[hidelinks]{hyperref}

\title{Adaptive Cross-Modal Fusion with Sparse Attention for Pedestrian Crossing Intention Prediction}

\author[1]{Md Mahfuzur Rahman}
\author[2]{Pengzhan Zhou}
\author[3]{A. F. M. Abdun Noor}
\author[4]{Md Imam Ahasan}
\author[5]{Kah Ong Michael Goh}
\author[6]{S. M. Hasan Mahmud}
\author[7]{Md Mustafizur Rahman}
\author[8]{Kaixin Gao}

\affil[1, 2, 4, 7, 8]{College of Computer Science, Chongqing University, Chongqing, China}
\affil[3, 6]{Department of Software Engineering, Daffodil International University, Bangladesh}
\affil[5]{Faculty of Information Science \& Technology, Multimedia University, Malaysia}

\corrauthor{Pengzhan Zhou, Kah Ong Michael Goh}{pzzhou@cqu.edu.cn, michael.goh@mmu.edu.my}

\begin{abstract}
Predicting pedestrian crossing intention is a safety-critical task for autonomous vehicles, yet existing methods rely on single-modal inputs or suboptimal fusion strategies that fail to model complementary visual and kinematic cues under varying traffic conditions. Most multimodal approaches apply dense cross-modal attention indiscriminately, introducing inter-modal noise from uninformative modality pairings and limiting deployment efficiency. We present ADAPT (Adaptive Domain-Aware Pedestrian crossing Transformer), a multimodal framework that jointly models local and global visual context alongside temporal kinematic dynamics over a 16-frame observation window at a 1-2 second time to event horizon. ADAPT processes four spatially registered image modalities alongside ego-vehicle speed, bounding boxes, and skeleton pose descriptors through five specialized modules: a weight-shared Swin Transformer V2 Tiny backbone for visual feature extraction, a Cross-Modality Guided Attention module that fuses local and global visual streams via hierarchical channel-spatial attention with a learnable adaptive routing gate, a two-layer Mamba State Space Model for linear-complexity kinematic encoding, a Sparse Cross-Modal Attention module that retains only top-$k$ inter-modal attention paths to suppress uninformative modality interactions, and a ViT-style Temporal Feature Fusion module for clip-level aggregation. On the JAAD benchmark, ADAPT achieves AUC of 0.73 on JAAD\textsubscript{beh} and 0.85 on JAAD\textsubscript{all}, surpassing all evaluated baselines. On PIE, it attains an accuracy of 0.92 and AUC of 0.90, exceeding the prior SOTA methods by up to 0.03 points. At 17.23\,ms per sample, ADAPT reduces inference latency by $2$-$4\times$ relative to architecturally comparable methods, establishing a strong balance between predictive accuracy and deployment efficiency.
\end{abstract}

\begin{document}

\flushbottom
\maketitle
\thispagestyle{empty}

\section{Introduction}

In recent years, advances in artificial intelligence and sensor technology have accelerated the development of autonomous vehicles (AV), especially in urban environments~\cite{chen2021interpretable}. Despite this progress, AVs still struggle to interact with other road users as effectively as experienced human drivers~\cite{crosato2022interaction}. One critical capability is predicting whether a pedestrian intends to cross the street. Accurate prediction allows AVs to anticipate pedestrian actions earlier and plan safer and smoother responses~\cite{rasouli2017agreeing}. However, predicting pedestrian intention remains challenging. Pedestrian behavior varies widely and depends on many contextual and social factors~\cite{rasouli2019autonomous, sharma2022pedestrian}. These factors include the surrounding environment, such as road layout and traffic conditions, as well as interactions with other road users. Early studies mainly relied on single-modal inputs, such as pedestrian images~\cite{rasouli2017they} or pose keypoints~\cite{fang2019intention}. However, using only one modality often leads to unreliable predictions because pedestrian behavior depends on multiple signals. For example, models that rely only on pedestrian images cannot capture how nearby vehicles or other traffic participants influence pedestrian decisions.

\begin{figure}[!ht]
    \centering
    \includegraphics[width=1\linewidth]{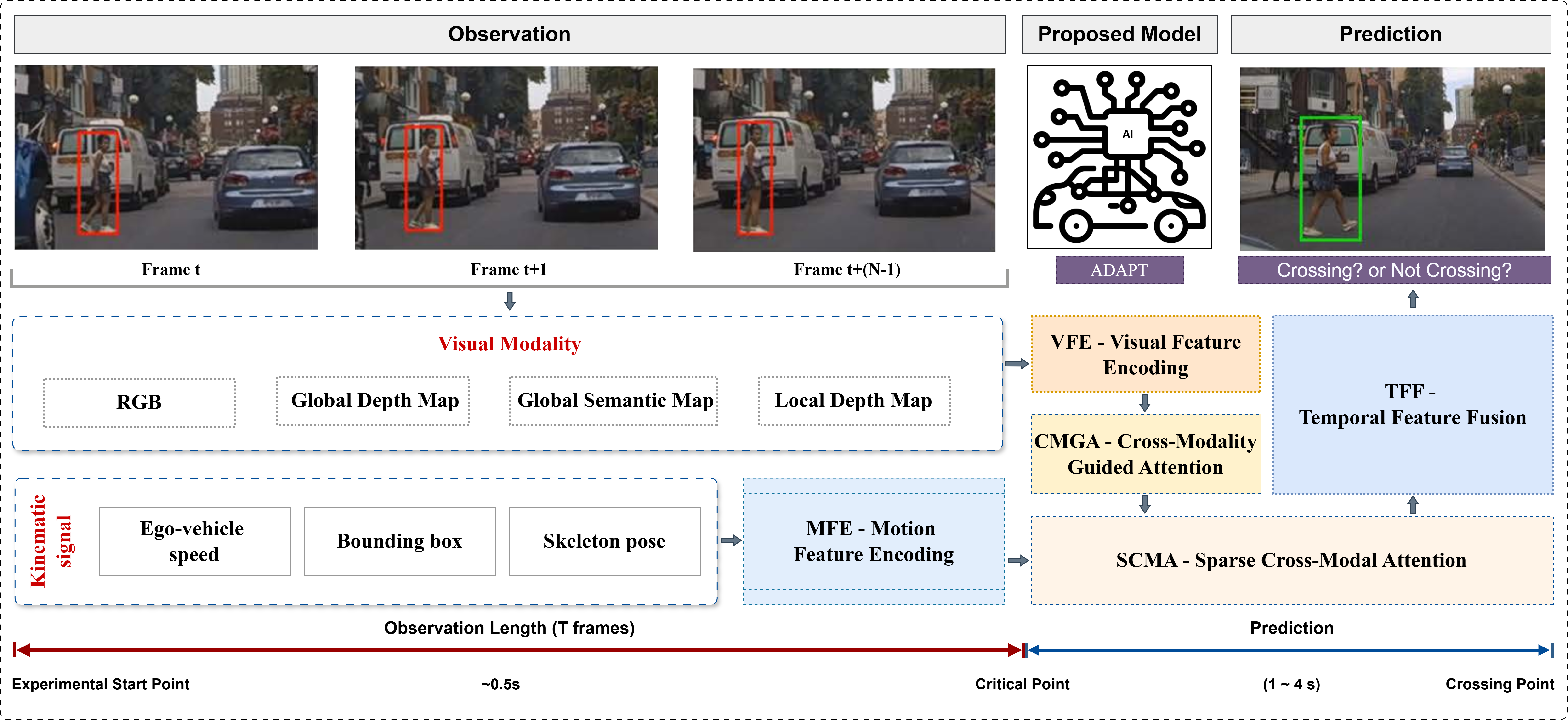}
    \caption{Overview of the proposed ADAPT framework for pedestrian crossing intention prediction. The model takes a sequence of observed frames and processes four spatially aligned visual modalities (RGB, global depth, global semantic map, and local depth) along with kinematic signals (ego-vehicle speed, pedestrian bounding boxes, and skeleton pose). Visual features are extracted via the Visual Feature Encoding (VFE) module and refined using Cross-Modality Guided Attention (CMGA). Motion dynamics are encoded by the Motion Feature Encoding (MFE) module. A Sparse Cross-Modal Attention (SCMA) module selectively integrates visual and motion representations, and the Temporal Feature Fusion (TFF) module aggregates temporal information to predict the crossing intention.}
    \label{fig:Motivation_concept}
\end{figure}

Recent work addresses this limitation by combining information from multiple modalities. Researchers have introduced additional visual inputs, such as semantic maps~\cite{yang2022predicting} and optical flow~\cite{bai2022deep}, as well as motion-related data like bounding boxes~\cite{kotseruba2021benchmark} and vehicle speed~\cite{rasouli2020pedestrian}. These modalities help capture both pedestrian-specific cues and the broader traffic context. Many studies process each modality with a separate branch that extracts features from the full observation sequence. The model then fuses these features to predict pedestrian intention. For example, Global PCPA~\cite{yang2022predicting, abdun2025geglunet} uses convolutional neural networks (CNNs) to extract visual features from image-based inputs. The model then combines these features with motion data and encodes them using separate gated recurrent units (GRUs) for temporal modeling. A modality-level attention module integrates the resulting features before prediction. Other approaches follow a similar strategy. In~\cite{yang2022predicting, wang2023pedestrian}, the transformers perform temporal modeling for each modality independently. The model then combines the features through multiscale stacking or transformer-based modality attention. Transformers have recently shown strong performance in sequential modeling and vision tasks, which has led to several Transformer-based multimodal fusion approaches. Action-ViT~\cite{zhao2021action} stacks visual modalities along the temporal dimension and converts nonvisual modalities into pseudo-images. A pretrained Vision Transformer processes these inputs to extract temporal features, followed by modality attention. In~\cite{wang2023pedestrian}, separate Transformers model driving behavior, traffic context, pedestrian actions, and visual information. The model then merges the features using multiscale stacking.

Other studies focus on modeling relationships across modalities. Lorenzo et al.~\cite{lorenzo2021intformer} and Osman et al.~\cite{osman2024multi} introduce Transformer-based modality attention to learn dependencies between modalities through attention scores. Inspired by neuroscience, PIT~\cite{zhou2023pit} proposes a progressive interaction framework that models dynamic relationships using Transformer attention and temporal fusion blocks. TAMformer~\cite{osman2023tamformer} introduces a multimodal Transformer architecture that learns attention masks to capture temporal dependencies. A later extension~\cite{osman2024multi} adds language inputs through modality distillation, using textual descriptions to improve contextual understanding. RAIDN~\cite{yang2024real} models pedestrian behavior and interactions with traffic participants using a multiscale graph Transformer combined with an interactive relational graph convolutional network. IntentFormer~\cite{sharma2025predicting} proposes a co-learning Transformer architecture that integrates multiple modalities through a multistage encoder with shared-weight attention, enabling hierarchical cross-modal interaction.

To address these limitations, we present \textbf{ADAPT} (Adaptive Domain-Aware Pedestrian crossing Transformer), a multimodal framework for pedestrian crossing intention prediction that jointly models local and global visual context alongside temporal kinematic dynamics. ADAPT processes four spatially registered image modalities RGB crops, local depth, global semantic segmentation, and global depth alongside three kinematic signals comprising ego-vehicle speed, pedestrian bounding boxes, and skeleton pose descriptors, over a 16-frame observation window at a time-to-event horizon of 1-2 seconds. Visual representations are extracted by a weight-shared Swin Transformer V2 Tiny backbone through a Visual Feature Encoding (VFE) module and refined through a Cross-Modality Guided Attention (CMGA) module, which applies hierarchical channel-spatial attention to fuse local and global visual streams via a learnable adaptive routing gate. Motion dynamics are encoded by a Motion Feature Encoding (MFE) module comprising a two-layer Mamba State Space Model, capturing long-range temporal kinematic dependencies with linear complexity. The resulting visual and motion representations are selectively integrated by a Sparse Cross-Modal Attention module (SCMA), which retains only the top-$k$ inter-modal attention paths to suppress noise from uninformative modality pairings. Finally, a ViT-style Temporal Feature Fusion module (TFF) aggregates frame-level representations over the full clip and produces the crossing-probability estimate. Extensive experiments on the JAAD and PIE benchmarks demonstrate that ADAPT achieves state-of-the-art accuracy while maintaining a competitive inference latency of 17.23\,ms per sample, establishing a favourable balance between representational capacity and deployment efficiency. The main contributions of this work are as follows.

\begin{itemize}
  \item
  We propose a hierarchical channel-spatial attention module that fuses complementary local (RGB + local depth) and global (semantic segmentation + global depth) visual streams. An adaptive routing gate dynamically weights the contribution of each stream on a per-sample basis, enabling the model to emphasize the most informative scene granularity under varying traffic conditions.
  \item
  We introduce a top-$k$ sparsity mask over pairwise inter-modal attention scores that selectively integrates visual and motion representations while discarding uninformative modality pairings. This principled sparsity constraint yields consistent gains over dense inter-modal attention across all evaluated benchmarks, confirming that selective fusion outperforms unrestricted attention when modalities carry heterogeneous information density.
  \item
  By pairing a lightweight Swin-V2-T backbone with a Mamba-based motion encoder, ADAPT achieves state-of-the-art accuracy on \textsc{jaad}\textsubscript{beh} (0.73) and PIE (0.92) while operating at 17.23\,ms per sample with a 2-4$\times$ latency reduction relative to architecturally comparable methods. Our results demonstrate that high predictive accuracy and deployment efficiency are simultaneously attainable in pedestrian intention prediction.
\end{itemize}

\section{Methodology}
\label{sec:method}

\subsection{Problem Formulation}
\label{sec:formulation}

\begin{figure*}[!ht]
    \centering
    \includegraphics[width=\textwidth]{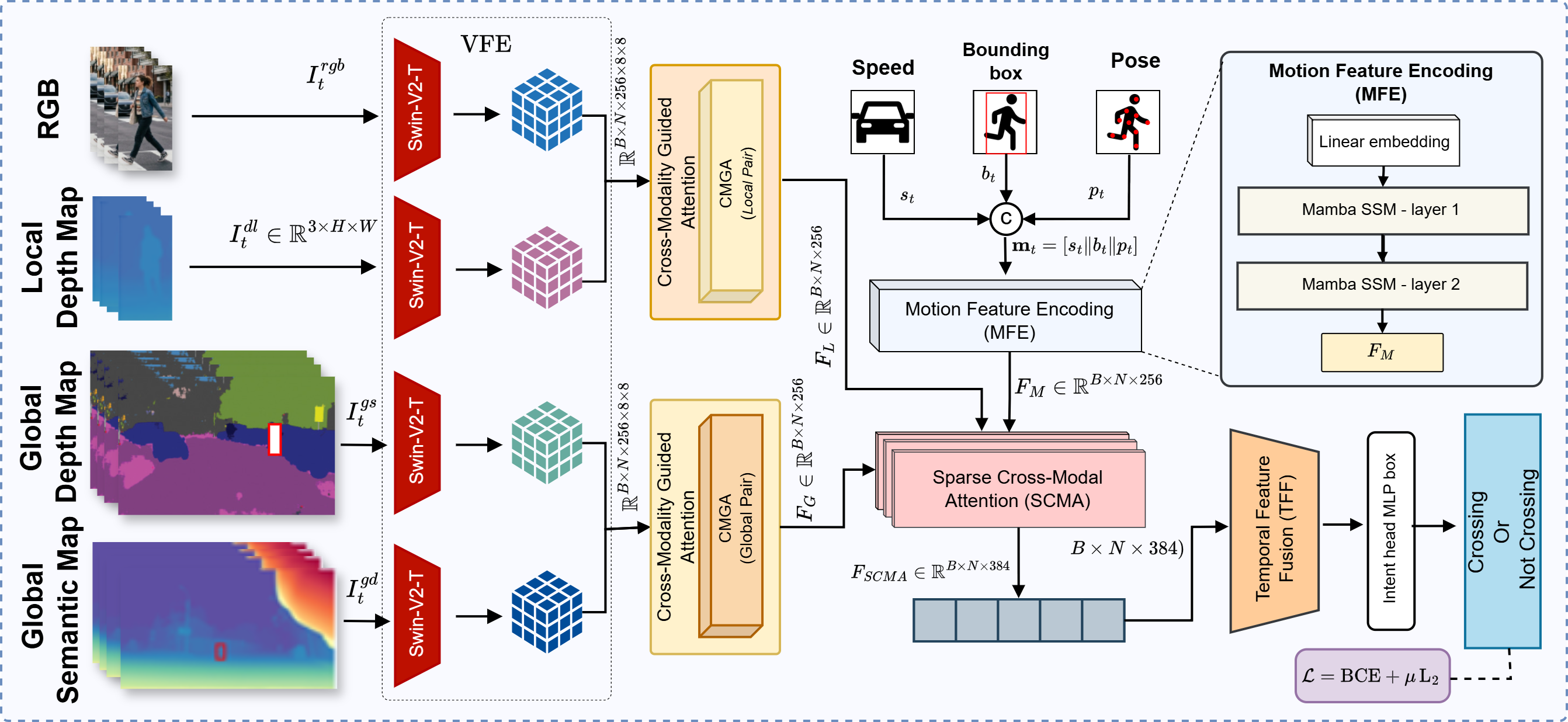}
        \caption{Architecture of ADAPT: VFE extracts per-modality spatial features, CMGA fuses local and global visual streams, MFE encodes motion dynamics via Mamba SSM, SCMA performs sparse cross-modal attention ($k{=}2$), and TFF aggregates temporal features via a ViT-style encoder to predict crossing probability $\hat{p}$.}
    \label{fig:adapt_architecture}
\end{figure*}

Let a pedestrian observation be represented as a clip of $N$ consecutive video frames extracted from a forward-facing onboard camera, where $N = 16$ in all experiments. Each frame $t \in \{1, \ldots, N\}$ contributes four spatially registered visual modalities: a local RGB crop $I^{\mathrm{rgb}}_{t} \in \mathbb{R}^{3 \times H \times W}$ centered on the pedestrian bounding box, a local depth map $I^{\mathrm{dl}}_{t} \in \mathbb{R}^{3 \times H \times W}$, a global semantic segmentation map $I^{\mathrm{gs}}_{t} \in \mathbb{R}^{3 \times H \times W}$, and a global depth map $I^{\mathrm{gd}}_{t} \in \mathbb{R}^{3 \times H \times W}$, where $H = W = 256$. Three kinematic signals are available per frame: ego-vehicle speed $s_t \in \mathbb{R}$, pedestrian bounding box coordinates $b_t \in \mathbb{R}^{4}$, and a body-pose descriptor $p_t \in \mathbb{R}^{36}$ encoding the two-dimensional positions and confidence scores of 17 skeletal keypoints. The crossing-intention prediction task is formulated as binary sequence classification. Given the multimodal clip observation
\begin{equation}
\mathcal{X} = \left\{ I^{\mathrm{rgb}}_{t},\, I^{\mathrm{dl}}_{t},\, I^{\mathrm{gs}}_{t},\, I^{\mathrm{gd}}_{t},\, s_t,\, b_t,\, p_t \right\}_{t=1}^{N},
\label{eq:input}
\end{equation}
the objective is to learn a mapping $f_\theta : \mathcal{X} \mapsto \hat{y} \in \{0, 1\}$, where $\hat{y} = 1$ denotes the pedestrian's intention to cross the road within a time-to-event (TTE) horizon of 1-2 seconds, and $\hat{y} = 0$ otherwise. The parameters $\theta$ are optimized on a labelled training set $\{(\mathcal{X}_i, y_i)\}_{i=1}^{M}$ under a composite loss described in Sec.~\ref{sec:loss}. The overall architecture, illustrated in Figure~\ref{fig:adapt_architecture}, processes $\mathcal{X}$ through a sequential pipeline of five modules. The Visual Feature Encoding module (VFE) extracts per-frame spatial representations from the four image modalities using a weight-shared hierarchical backbone. The Motion Feature Encoding module (MFE) encodes temporal kinematic dynamics from the concatenated scalar and vector motion signals. The Cross-Modality Guided Attention module (CMGA) fuses local and global visual streams through hierarchical channel-spatial attention with an adaptive routing gate, while the Sparse Cross-Modal Attention module (SCMA) selectively integrates visual and motion representations. Finally, the Temporal Feature Fusion module (TFF) aggregates frame-level features over the full clip to produce a scalar crossing-probability estimate.

\subsection{Visual Feature Encoding}
\label{sec:vfe}
The VFE module extracts a shared spatial representation from each of the four image modalities independently. A single Swin Transformer V2 Tiny (Swin-V2-T) backbone~\cite{liu2022swin} is shared across all four modalities and configured with image resolution $256 \times 256$, window size 16, embedding dimension 96, depth sequence $[2, 2, 6, 2]$, and head counts $[3, 6, 12, 24]$ across the four hierarchical stages. The final stage produces feature channels of dimension $C_{\mathrm{swin}} = 96 \times 2^{3} = 768$, yielding approximately 28 million backbone parameters, a 68\% reduction relative to the Swin-V2-B configuration (embed\_dim $= 128$, depths $[2, 2, 18, 2]$, ${\approx}88\text{M}$ parameters), achieved primarily by reducing the stage-3 block count from 18 to 6 and lowering the initial embedding dimension from 128 to 96. The backbone is initialized from SimMIM~\cite{xie2022simmim} pretrained weights. The self-supervised masked image modelling objective encourages the backbone to capture generalizable spatial structure rather than label-discriminative features, improving robustness to the diverse illumination and scene conditions present in both JAAD and PIE. For a single modality input $I_t \in \mathbb{R}^{3 \times 256 \times 256}$, the Swin-V2-T feature extractor produces a patch token sequence $Z_t \in \mathbb{R}^{L \times C_{\mathrm{swin}}}$, where $L = 64$ denotes the spatial token count. These tokens are reshaped into a feature map $\hat{Z}_t \in \mathbb{R}^{C_{\mathrm{swin}} \times \sqrt{L} \times \sqrt{L}}$ and projected to a lower-dimensional representation via a $1 \times 1$ convolution followed by batch normalization and ReLU activation:
\begin{equation}
F_t = \mathrm{BN}\!\left(\mathrm{Conv}_{1\times1}(\hat{Z}_t)\right) \in \mathbb{R}^{256 \times 8 \times 8},
\label{eq:vfe_proj}
\end{equation}
where the projection reduces the channel dimension from 768 to 256 while preserving the $8 \times 8$ spatial grid structure. This shared output dimensionality ensures that all downstream modules operate identically regardless of the backbone variant employed. Processing all $N$ frames in a batch is accomplished by merging the batch and temporal axes prior to backbone inference. For a modality tensor $I \in \mathbb{R}^{B \times N \times 3 \times 256 \times 256}$, the input is reshaped to $(BN) \times 3 \times 256 \times 256$, processed through Eq.~(\ref{eq:vfe_proj}), and subsequently restored to $\mathbb{R}^{B \times N \times 256 \times 8 \times 8}$. Applying this procedure to all four modalities yields the per-frame feature map set
\begin{equation}
\left\{ F^{\mathrm{rgb}},\, F^{\mathrm{dl}},\, F^{\mathrm{gs}},\, F^{\mathrm{gd}} \right\}, \quad F^{(\cdot)} \in \mathbb{R}^{B \times N \times 256 \times 8 \times 8},
\label{eq:vfe_out}
\end{equation}
which serve as inputs to the subsequent cross-modal fusion stages. To stabilize training, the patch embedding layer and the first two Swin-V2-T hierarchical stages are frozen throughout all training phases, while the higher-level stages are progressively unfrozen according to the multi-stage schedule described.

\subsection{Motion Feature Encoding}
\label{sec:mfe}
The MFE module encodes the temporal dynamics of pedestrian kinematics and ego-vehicle motion from the per-frame motion signals. For each frame $t$, the speed scalar $s_t \in \mathbb{R}$, bounding box vector $b_t \in \mathbb{R}^{4}$, and pose descriptor $p_t \in \mathbb{R}^{36}$ are concatenated to form a joint motion vector $m_t \in \mathbb{R}^{41}$. The scalar speed channel is normalized to $[0, 1]$ by dividing by the dataset-specific maximum speed (60 km/h for JAAD, 70 km/h for PIE), and the bounding box coordinates are normalized by the image spatial dimension $H = W = 256$. Over the clip, the full motion sequence is $M \in \mathbb{R}^{B \times N \times 41}$. The sequence $M$ is projected into a $d_m = 256$ dimensional latent space via a linear embedding layer followed by layer normalization:
\begin{equation}
E = \mathrm{LayerNorm}(M W_e), \quad W_e \in \mathbb{R}^{41 \times 256}, \quad E \in \mathbb{R}^{B \times N \times 256}.
\label{eq:mfe_embed}
\end{equation}

Temporal modelling is performed by a two-layer Mamba State Space Model (SSM)~\cite{gu2024mamba}. Each Mamba layer operates with a selective scan mechanism that parameterizes the hidden state transition matrices $A$, $B$, $C$, and the feedthrough operator $D$ as functions of the input, enabling the model to selectively propagate or suppress information across time steps. This selective mechanism captures long-range dependencies between ego-vehicle deceleration cues and pedestrian bounding box displacement trajectories while maintaining linear $\mathcal{O}(N)$ time complexity with respect to sequence length, in contrast to the $\mathcal{O}(N^2)$ complexity of a comparable Transformer encoder. Each Mamba layer is configured with state dimension $d_{\mathrm{state}} = 16$, convolutional kernel size $d_{\mathrm{conv}} = 4$, and expansion factor $E = 2$. Letting $\mathrm{Mamba}_l(\cdot)$ denote the $l$-th Mamba layer, the two-layer encoding with residual connections is expressed as
\begin{equation}
H^{(l)} = \mathrm{Mamba}_l\!\left(H^{(l-1)}\right) + H^{(l-1)}, \quad l \in \{1, 2\}, \quad H^{(0)} = E,
\label{eq:mamba_residual}
\end{equation}
and the final motion feature sequence is produced by layer normalization of the output of the second layer:
\begin{equation}
F_M = \mathrm{LayerNorm}\!\left(H^{(2)}\right) \in \mathbb{R}^{B \times N \times 256}.
\label{eq:mfe_out}
\end{equation}
The motion feature tensor $F_M$ encodes the complete temporal trajectory of the kinematic state and is subsequently fused with visual representations in the SCMA module.

\subsection{Cross-Modal Fusion: CMGA and SCMA}
\label{sec:fusion}

\textbf{Cross-Modality Guided Attention:} The CMGA module fuses the four visual feature maps produced by the VFE into two compact frame-level descriptors: a local visual feature $F_L \in \mathbb{R}^{B \times N \times 256}$ and a global visual feature $F_G \in \mathbb{R}^{B \times N \times 256}$, through a hierarchical channel-spatial attention mechanism augmented with an adaptive routing gate. The fusion proceeds through two parallel pathways operating on complementary scene granularities. The local pathway processes the pair $(F^{\mathrm{rgb}}, F^{\mathrm{dl}})$, which encode pedestrian appearance and foreground depth, while the global pathway processes the pair $(F^{\mathrm{gs}}, F^{\mathrm{gd}})$, which encode scene-level semantic context and background geometry. Within each pathway, a CMGAPair block applies cross-modality guided channel attention (CMGCA), guided spatial attention (CMGSA), and adaptive fusion (CMGAF) sequentially. The CMGCA sub-module produces a channel-wise attention map that recalibrates feature $F_i$ conditioned on the complementary modality $F_j$. Given two input maps $F_i, F_j \in \mathbb{R}^{C \times H' \times W'}$ where $C = 256$ and $H' = W' = 8$, the channel descriptor for the guide $F_j$ is formed by combining global average pooling and global max pooling:
\begin{equation}
d_j = \frac{1}{H'W'} \sum_{h,w} F_j^{h,w} + \max_{h,w} F_j^{h,w} \;\in \mathbb{R}^{C}.
\label{eq:cmgca_desc}
\end{equation}
This descriptor is passed through a two-layer MLP with intermediate dimension $C/r$ (reduction ratio $r = 16$) and sigmoid activation to produce a channel weight vector $\alpha_j \in (0, 1)^C$, and the attended feature map is
\begin{equation}
\tilde{F}^{\mathrm{ch}}_i = F_i \odot \alpha_j,
\label{eq:cmgca_attend}
\end{equation}
where $\odot$ denotes element-wise multiplication broadcast over spatial dimensions. The CMGCA operation is applied in both directions, yielding $\tilde{F}^{\mathrm{ch}}_i$ and $\tilde{F}^{\mathrm{ch}}_j$. The CMGSA sub-module subsequently applies a spatially selective mask to each channel-attended feature, conditioned on the complementary modality. For input $\tilde{F}^{\mathrm{ch}}_i$ guided by $\tilde{F}^{\mathrm{ch}}_j$, a two-channel spatial descriptor is formed by concatenating the channel-mean and channel-max projections of the guide:
\begin{equation}
D^{\mathrm{sp}}_j = \left[ \mathrm{AvgPool}_C\!\left(\tilde{F}^{\mathrm{ch}}_j\right) \;\|\; \mathrm{MaxPool}_C\!\left(\tilde{F}^{\mathrm{ch}}_j\right) \right] \in \mathbb{R}^{2 \times H' \times W'},
\label{eq:cmgsa_desc}
\end{equation}
where $\|$ denotes channel concatenation. A $3 \times 3$ convolutional layer with sigmoid activation maps $D^{\mathrm{sp}}_j$ to a spatial attention map $\beta_j \in (0, 1)^{H' \times W'}$, and the spatially attended feature is computed as
\begin{equation}
\tilde{F}^{\mathrm{sp}}_i = \tilde{F}^{\mathrm{ch}}_i \odot \beta_j + F_i,
\label{eq:cmgsa_attend}
\end{equation}
where the residual term $F_i$ preserves gradient flow. The CMGAF sub-module then produces the fused pair feature map via additive and multiplicative fusion paths:
\begin{equation}
F^{\mathrm{pair}} = \left[ \mathrm{Conv}\!\left(\tilde{F}^{\mathrm{sp}}_i + \tilde{F}^{\mathrm{sp}}_j\right) \;\|\; \mathrm{Conv}\!\left(\tilde{F}^{\mathrm{sp}}_i \odot \tilde{F}^{\mathrm{sp}}_j\right) \right] \in \mathbb{R}^{256 \times H' \times W'},
\label{eq:cmgaf}
\end{equation}
where each $\mathrm{Conv}$ is a $3 \times 3$ convolution followed by batch normalization and ReLU, reducing the channel dimension to 128 before concatenation. Applying CMGAPair independently to the local and global streams yields feature maps $F^{\mathrm{local}}_{\mathrm{map}}, F^{\mathrm{global}}_{\mathrm{map}} \in \mathbb{R}^{(BN) \times 256 \times 8 \times 8}$. Global average pooling collapses the spatial dimensions:
\begin{equation}
f_\ell = \frac{1}{H'W'} \sum_{h,w} F^{\mathrm{local}}_{\mathrm{map}}\big|^{h,w} \in \mathbb{R}^{BN \times 256}, \quad
f_g = \frac{1}{H'W'} \sum_{h,w} F^{\mathrm{global}}_{\mathrm{map}}\big|^{h,w} \in \mathbb{R}^{BN \times 256}.
\label{eq:gap}
\end{equation}
An adaptive routing gate $g \in \mathbb{R}^{2}$ is computed from the concatenated pooled descriptors via a linear layer followed by softmax:
\begin{equation}
g = \mathrm{softmax}\!\left(W_g \left[f_\ell \;\|\; f_g\right]\right), \quad W_g \in \mathbb{R}^{2 \times 512},
\label{eq:gate}
\end{equation}
where $g = [g_1, g_2]^\top$ with $g_1 + g_2 = 1$. The gate dynamically modulates the contribution of each stream on a per-sample basis. The output local feature $F_L$ is produced by a weighted combination followed by a linear projection:
\begin{equation}
F_L = W_o\!\left(g_1 \cdot f_\ell + g_2 \cdot f_g\right) \in \mathbb{R}^{B \times N \times 256}, \quad W_o \in \mathbb{R}^{256 \times 256}.
\label{eq:fl}
\end{equation}
The global visual feature is computed as an unweighted additive fusion:
\begin{equation}
F_G = f_\ell + f_g \in \mathbb{R}^{B \times N \times 256}.
\label{eq:fg}
\end{equation}

\textbf{Sparse Cross-Modal Attention:} The SCMA module selectively integrates the three per-frame feature streams $F_L, F_G \in \mathbb{R}^{B \times N \times 256}$ and $F_M \in \mathbb{R}^{B \times N \times 256}$ into a unified cross-modal representation. Let $\{F_1, F_2, F_3\} = \{F_L, F_G, F_M\}$. For each stream $i$, dedicated projection matrices $W^Q_i, W^K_i, W^V_i \in \mathbb{R}^{256 \times d_p}$ with $d_p = 128$ produce query, key, and value representations:
\begin{equation}
Q_i = F_i W^Q_i, \quad K_i = F_i W^K_i, \quad V_i = F_i W^V_i, \quad Q_i, K_i, V_i \in \mathbb{R}^{B \times N \times d_p}.
\label{eq:qkv}
\end{equation}
A $3 \times 3$ pairwise score matrix $S \in \mathbb{R}^{B \times N \times 3 \times 3}$ is constructed by computing the scaled inner product between each query-key pair:
\begin{equation}
S_{ij} = \frac{Q_i \cdot K_j^\top}{\sqrt{d_p}}, \quad i, j \in \{1, 2, 3\}.
\label{eq:scores}
\end{equation}
A binary sparsity mask $M^{\mathrm{sp}} \in \{0, 1\}^{B \times N \times 3 \times 3}$ is derived by retaining only the top-$k$ entries in $S$ along the key dimension for each query row ($k = 2$):
\begin{equation}
M^{\mathrm{sp}}_{ij} =
\begin{cases}
1 & \text{if } j \in \mathrm{top}\text{-}k\!\left(\{S_{ij}\}_{j=1}^{3}\right), \\
0 & \text{otherwise.}
\end{cases}
\label{eq:mask}
\end{equation}
Scores at masked positions are set to $-\infty$ before softmax normalization:
\begin{equation}
A = \mathrm{softmax}\!\left(S \odot M^{\mathrm{sp}} + (1 - M^{\mathrm{sp}}) \cdot (-\infty)\right) \in \mathbb{R}^{B \times N \times 3 \times 3}.
\label{eq:sparse_attn}
\end{equation}
The attended value representations for each stream are obtained by weighting the stacked value matrix $V_{\mathrm{stack}} \in \mathbb{R}^{B \times N \times 3 \times d_p}$:
\begin{equation}
\tilde{V}_i = \sum_{j=1}^{3} A_{ij} V_j \in \mathbb{R}^{B \times N \times d_p},
\label{eq:attended_v}
\end{equation}
and the three attended representations are concatenated to form the SCMA output:
\begin{equation}
F_{\mathrm{SCMA}} = \left[\tilde{V}_1 \;\|\; \tilde{V}_2 \;\|\; \tilde{V}_3\right] \in \mathbb{R}^{B \times N \times 3d_p},
\label{eq:scma_out}
\end{equation}
where $3d_p = 384$. The sparsity mask $M^{\mathrm{sp}}$ is retained as a model output for attention visualization and interpretability analysis. By restricting each query stream to attend over only its two highest-scoring key streams, SCMA eliminates attention paths that carry low mutual information between modalities, a property empirically validated through the ablation study in Sec.~\ref{sec:Ablation}.

\subsection{Temporal Feature Fusion}
\label{sec:tff}

The TFF module aggregates the per-frame cross-modal representations $F_{\mathrm{SCMA}} \in \mathbb{R}^{B \times N \times 384}$ over the clip dimension through a ViT-style Transformer encoder and produces a scalar crossing-probability estimate. A learnable classification token $e_{\mathrm{cls}} \in \mathbb{R}^{1 \times 384}$ is prepended to the frame sequence, and a learnable positional embedding $P \in \mathbb{R}^{(N+1) \times 384}$ is added to encode frame order:
\begin{equation}
X^{(0)} = \left[e_{\mathrm{cls}} \;\|\; F_{\mathrm{SCMA}}\right] + P \in \mathbb{R}^{B \times (N+1) \times 384}.
\label{eq:tff_input}
\end{equation}
Both $e_{\mathrm{cls}}$ and $P$ are initialized by truncated normal sampling with standard deviation 0.02. The sequence $X^{(0)}$ is processed by a four-layer Transformer encoder with eight attention heads, MLP expansion ratio 4, and dropout rate 0.1, followed by layer normalization:
\begin{equation}
X^{(L)} = \mathrm{LayerNorm}\!\left(\mathrm{TransEnc}\!\left(X^{(0)}\right)\right), \quad L = 4.
\label{eq:tff_enc}
\end{equation}
The CLS token aggregates global temporal context, and the final crossing probability is obtained by passing the CLS representation through a two-layer MLP classifier:
\begin{equation}
\hat{p} = \sigma\!\left(W_2\, \mathrm{ReLU}\!\left(W_1\, X^{(L)}_{[:,0,:]} + b_1\right) + b_2\right),
\label{eq:classifier}
\end{equation}
where $W_1 \in \mathbb{R}^{128 \times 384}$, $W_2 \in \mathbb{R}^{1 \times 128}$, dropout with rate 0.3 is applied between the two linear layers, and $\sigma(\cdot)$ denotes the sigmoid function. The final prediction is $\hat{y} = \mathbf{1}[\hat{p} \geq 0.5]$.

\subsection{Loss Function and Training Protocol}
\label{sec:loss}

\textbf{Loss Function:} Training is performed under a composite loss
\begin{equation}
\mathcal{L} = \mathcal{L}_{\mathrm{intent}} + \mu\, \mathcal{L}_{\mathrm{reg}},
\label{eq:loss_total}
\end{equation}
where $\mu = 10^{-3}$ controls the regularization strength. The primary term $\mathcal{L}_{\mathrm{intent}}$ is a class-weighted binary cross-entropy:
\begin{equation}
\mathcal{L}_{\mathrm{intent}} = -\frac{1}{B} \sum_{i=1}^{B} w_i \left[ y_i \log \hat{p}_i + (1 - y_i) \log (1 - \hat{p}_i) \right],
\label{eq:loss_intent}
\end{equation}
where $w_i$ is the per-sample class weight assigned according to the inverse-frequency scheme $w_c = M / (2 n_c)$, with $M$ the total training sample count and $n_c$ the count of samples in class $c \in \{0, 1\}$. The regularization term $\mathcal{L}_{\mathrm{reg}}$ is the squared Frobenius norm of the weight matrices in the intent head MLP:
\begin{equation}
\mathcal{L}_{\mathrm{reg}} = \sum_{W \in \Theta_{\mathrm{head}}} \|W\|^2_F,
\label{eq:loss_reg}
\end{equation}
where $\Theta_{\mathrm{head}}$ denotes the set of weight matrices (excluding biases) in the TFF classifier head.

\textbf{Training Protocol:} The model is trained using a three-stage progressive fine-tuning protocol. In Stage 1 (epochs 1-20), the entire VFE backbone is frozen; only the MFE, CMGA gate, SCMA, and TFF parameters are updated, with the Adam optimizer at learning rate $2 \times 10^{-5}$ and zero backbone learning rate. In Stage 2 (epochs 21-80), the last two Swin-V2-T hierarchical stages are unfrozen with a reduced backbone learning rate of $2 \times 10^{-6}$, while the remaining modules retain their Stage 1 learning rate. In Stage 3 (epochs 81-150), all backbone stages are unfrozen, and at epoch 100 all learning rates are decayed by a factor of 0.1. Gradient norms are clipped to a maximum of 1.0 throughout all stages. Training employs FP16 automatic mixed precision (AMP) with PyTorch's GradScaler and is distributed across four GPUs using DistributedDataParallel (DDP) with SyncBatchNorm. The total batch size is 32 (8 per GPU). A WeightedRandomSampler is applied in non-distributed settings to counteract class imbalance at the sampling level, complementing the per-sample loss weighting in Eq.~(\ref{eq:loss_intent}). Early stopping with patience 25 epochs is applied based on the validation AUC metric. The JAAD and PIE datasets are trained and evaluated independently, with no cross-dataset knowledge transfer.

\begin{table*}[!ht]
\centering
\small
\setlength{\tabcolsep}{4pt}
\renewcommand{\arraystretch}{1.12}
\caption{Quantitative evaluation of ADAPT against existing pedestrian intention
prediction methods on JAAD$_{beh}$ and JAAD$_{all}$ splits. Bold denotes the
best result per column.}
\label{tab:jaad_comparison}

\begin{threeparttable}
\begin{tabular}{lcccccccccc}
\toprule
\multirow{2}{*}{\textbf{Models}} 
& \multicolumn{5}{c}{\textbf{JAAD$_{beh}$}} 
& \multicolumn{5}{c}{\textbf{JAAD$_{all}$}} \\
\cmidrule(lr){2-6} \cmidrule(lr){7-11}
& \textbf{Acc} & \textbf{AUC} & \textbf{F1} & \textbf{Precision} & \textbf{Recall}
& \textbf{Acc} & \textbf{AUC} & \textbf{F1} & \textbf{Precision} & \textbf{Recall} \\
\midrule
SF-GRU~\cite{rasouli2020pedestrian}              & 0.58 & 0.56 & 0.65 & 0.68 & 0.62 & 0.76 & 0.77 & 0.53 & 0.40 & 0.79 \\
FUSSI~\cite{piccoli2020fussi}               & 0.59 & 0.58 & 0.69 & 0.66 & 0.73 & 0.60 & 0.72 & 0.40 & 0.27 & 0.73 \\
SingleRNN~\cite{kotseruba2020they}           & 0.60 & 0.54 & 0.70 & 0.65 & 0.76 & 0.78 & 0.77 & 0.54 & 0.42 & 0.75 \\
PCPA~\cite{kotseruba2021benchmark}                & 0.56 & 0.54 & 0.63 & 0.66 & 0.60 & 0.77 & 0.79 & 0.52 & 0.42 & 0.83 \\
Global PCPA~\cite{yang2022predicting}         & 0.62 & 0.54 & 0.74 & 0.65 & 0.85 & 0.83 & 0.82 & 0.63 & 0.51 & 0.81 \\
V-PedCross~\cite{bai2022deep}           & 0.61 & 0.50 & 0.75 & 0.71 & 0.80 & 0.82 & 0.74 & 0.64 & 0.58 & 0.63 \\
MMTN~\cite{wang2023pedestrian}                & 0.68 & 0.63 & 0.76 & 0.71 & 0.81 & 0.89 & 0.78 & 0.66 & 0.72 & 0.61 \\
PIT~\cite{zhou2023pit}                 & 0.70 & 0.65 & 0.81 & 0.71 & 0.93 & 0.87 & 0.87 & 0.66 & 0.54 & 0.85 \\
STTF-MANet~\cite{zhang2024multi}          & 0.66 & 0.58 & 0.77 & 0.67 & 0.89 & 0.89 & 0.80 & 0.67 & 0.68 & 0.67 \\
RAIDN~\cite{yang2024real}               & 0.70 & 0.69 & 0.75 & 0.73 & 0.78 & 0.89 & 0.80 & 0.66 & 0.65 & 0.72 \\
MTC~\cite{li2025mtc} & 0.71 & 0.65 & 0.80 & 0.72 & 0.90 & 0.90 & 0.82 & 0.70 & 0.70 & 0.70 \\
\midrule
\textbf{ADAPT (our)}             & 0.74 & 0.70 & 0.83 & 0.76 & 0.92 & 0.91 & 0.85 & 0.76 & 0.74 & 0.78 \\
\bottomrule
\end{tabular}
\end{threeparttable}
\end{table*}

\section{Experimental Setup}

\label{sec:experiments}

\subsection{Datasets}
\label{subsec:datasets}
Experiments are conducted on two publicly available benchmarks for pedestrian crossing intention prediction: 
the Joint Attention in Autonomous Driving dataset (JAAD)\footnote{JAAD dataset. DOI: \url{https://doi.org/10.48550/arXiv.1609.04741}. Website: \url{https://data.nvision2.eecs.yorku.ca/JAAD_dataset/}.}~\cite{rasouli2017they} 
and the Pedestrian Intention Estimation dataset (PIE)\footnote{PIE dataset. DOI: \url{https://doi.org/10.1109/ICCV.2019.00636}. Website: \url{https://data.nvision2.eecs.yorku.ca/PIE_dataset/}.}~\cite{rasouli2019pie}. 
Each dataset is treated as an independent experimental setting, with ADAPT trained and evaluated separately on each to ensure comparability with prior work under established benchmark protocols.

\textbf{JAAD:} The JAAD dataset comprises 346 video sequences recorded from a forward-facing monocular camera mounted on an ego-vehicle navigating urban intersections. It provides per-pedestrian bounding box annotations, crossing behaviour labels, and contextual attributes including traffic signal state, occlusion level, and pedestrian motion direction. The dataset is partitioned following the benchmark protocol established by Kotseruba~et~al.~\cite{kotseruba2021benchmark}, which allocates 177 videos for training, 29 for validation, and 117 for testing. Two evaluation subsets are defined: \textsc{JAADbeh}, comprising pedestrians annotated with explicit behaviour labels, and \textsc{JAADall}, encompassing all annotated pedestrians regardless of behavioural annotation availability. Following the MTC experimental protocol~\cite{li2025mtc}, observation clips are extracted with a fixed temporal window of $N = 16$ frames at a temporal overlap ratio of $80\%$, corresponding to a stride of $\lfloor N(1 - 0.8) \rfloor = 3$ frames. The observation window spans $0.5$~seconds of ego-vehicle motion, and clips are retained only when the time-to-event (TTE) satisfies $1\,\text{s} \leq \text{TTE} \leq 2\,\text{s}$, ensuring that predictions are made sufficiently in advance of the pedestrian's crossing action to be actionable for downstream vehicle control. Ego-vehicle speed is normalised to the unit interval $[0, 1]$ using a dataset-specific upper bound of $60$~km/h, and bounding box coordinates are normalised by the image dimensions $H = W = 256$ pixels to produce scale-invariant representations.

\textbf{PIE:} The PIE dataset consists of continuous driving sequences recorded across six data collection sets (\texttt{set01}-\texttt{set06}) under diverse traffic conditions, including varying illumination, ego-vehicle speed profiles, and pedestrian behaviours. Unlike JAAD, PIE provides dense ego-motion telemetry synchronised with video frames, offering richer ego-speed signal. The dataset split follows the protocol adopted by the MFT baseline~\cite{li2025multi}: sets \texttt{set01}, \texttt{set02}, and \texttt{set06} constitute the training partition; sets \texttt{set04} and \texttt{set05} serve as the validation partition; and \texttt{set03} constitutes the held-out test partition. Clip extraction applies the same temporal configuration as JAAD ($N=16$, $80\%$ overlap, TTE $\in [1, 2]$~s). Ego-vehicle speed is normalised using an upper bound of $70$~km/h, reflecting the higher velocity range observed in PIE's continuous-driving scenarios compared to JAAD's intersection-focused recordings.

\begin{figure}[!ht]
    \centering
    \includegraphics[width=1\linewidth]{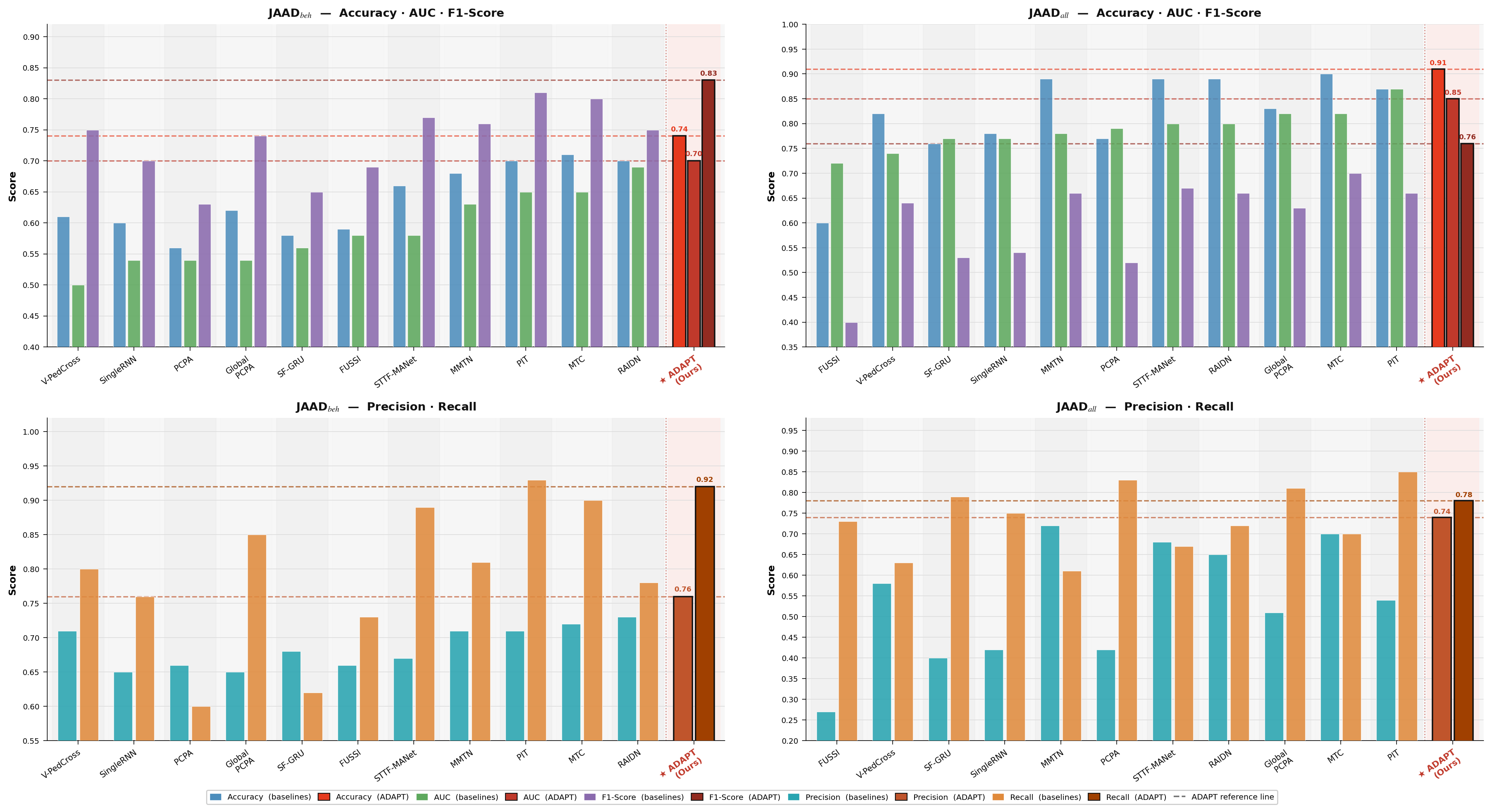}
    \caption{Comparison of ADAPT with state-of-the-art methods on
    (a)~JAAD$_\text{beh}$ and (b)~JAAD$_\text{all}$.
    Methods sorted by JAAD$_\text{beh}$ AUC (ascending).
    Dashed lines mark ADAPT's scores per metric.
    Best results in bold; `-' indicates unreported values.}
    \label{fig:jaad_comparison_chart}
\end{figure}

\subsection{Implementation Details}
\label{sec:impl}
All experiments are conducted independently on two publicly available benchmarks  the Joint Attention in Autonomous Driving (JAAD)~\cite{rasouli2017they} dataset and the Pedestrian Intention Estimation (PIE)~\cite{rasouli2019pie} dataset  with no cross-dataset training or knowledge transfer between the two. Following the protocol of~\cite{kotseruba2021benchmark}, JAAD is partitioned into 177 training, 29 validation, and 117 test video sequences, while PIE is split by recording session, with \texttt{set01}, \texttt{set02}, and \texttt{set06} for training, \texttt{set04} and \texttt{set05} for validation, and \texttt{set03} for evaluation; in both cases, clips are extracted with a 16-frame observation window, 80\% temporal overlap, and a time-to-event (TTE) constraint of 1-2 seconds. All visual inputs are resized to $256 \times 256$ pixels; RGB and semantic frames are normalized to $[0,1]$ and standardized with ImageNet channel statistics ($\boldsymbol{\mu} = [0.485, 0.456, 0.406]$, $\boldsymbol{\sigma} = [0.229, 0.224, 0.225]$), single-channel depth maps are replicated across three channels and normalized identically, bounding box coordinates are divided by the image spatial dimension to yield values in $[0,1]$, and ego-vehicle speed is clipped and normalized by the dataset-specific maximum (60~km/h for JAAD; 70~km/h for PIE). During training, a consistent random horizontal flip with probability 0.5 is applied simultaneously across all four visual modality sequences within a clip to preserve spatial registration; no augmentation is used at validation or test time. The proposed model is implemented in PyTorch~\cite{imambi2021pytorch} ($\geq$2.1.0) with the timm library ($\geq$0.9.12) supplying the Swin Transformer V2 backbone and the mamba-ssm package ($\geq$1.2.0) providing the selective state space model layers. Training is distributed across four NVIDIA GPUs via DistributedDataParallel (DDP) with NCCL backend and SyncBatchNorm, with FP16 automatic mixed precision (AMP) enabled throughout. The total batch size is 32 (8 per GPU), and the Adam optimizer~\cite{zhang2018improved} is employed with weight decay set to zero and gradient norms clipped to 1.0. The VFE backbone is initialized from a SimMIM-pretrained Swin-V2-B checkpoint; all remaining modules use PyTorch default initialization. Training proceeds for up to 150 epochs under the three-stage progressive fine-tuning schedule described, with early stopping applied after 25 consecutive epochs without improvement in validation AUC; the checkpoint attaining the highest validation AUC is selected for final test evaluation. Class imbalance is addressed jointly through a WeightedRandomSampler at the data loading level and through per-sample inverse-frequency weights within the loss function. The regularization coefficient is fixed at $\mu = 10^{-3}$, the SCMA sparsity parameter at $k = 2$, the MFE Mamba encoder at 2 layers with state dimension $d_{\mathrm{state}} = 16$, convolutional kernel $d_{\mathrm{conv}} = 4$, and expansion factor $E = 2$, and the TFF Transformer at 4 layers with 8 attention heads, MLP expansion ratio 4, and classifier dropout rate 0.3.

\begin{table}[!ht]
\centering
\caption{Quantitative evaluation of ADAPT against existing pedestrian intention
prediction methods on the PIE dataset. Bold denotes the best result per column.}
\label{tab:PIE_comparison}
\begin{tabular}{lccccc}
\toprule
Models & Acc & AUC & F1 & Precision & Recall \\
\midrule
PCPA~\cite{kotseruba2021benchmark}  & 0.87 & 0.85 & 0.78 & 0.76 & 0.81 \\
Global PCPA~\cite{yang2022predicting} & 0.89 & 0.86 & 0.80 & 0.79 & 0.81 \\
TrouSPI-Net~\cite{gesnouin2021trouspi} & 0.88 & 0.88 & 0.80 & 0.73 & 0.89 \\
V-PedCross~\cite{bai2022deep} & 0.89 & 0.88 & 0.67 & 0.74 & 0.84 \\
STTF-MANet~\cite{zhang2024multi} & 0.89 & 0.88 & 0.82 & 0.79 & 0.85 \\
MTMGN~\cite{yang2024explainable} & 0.90 & 0.87 & 0.92 & 0.95 & 0.90 \\
PFRN~\cite{lv2024pedestrian} & 0.90 & 0.85 & 0.77 & 0.81 & 0.74 \\
Dual-STGAT~\cite{lian2025dual} & 0.86 & 0.87 & 0.91 & 0.92 & 0.90 \\
LSOP-Net~\cite{liu2025long} & 0.89 & 0.87 & 0.81 & 0.80 & 0.82 \\
\midrule
\textbf{ADAPT (Ours)} & 0.92 & 0.90 & 0.83 & 0.84 & 0.81 \\
\bottomrule
\end{tabular}
\end{table}

\subsection{Evaluation Metrics}
\label{sec:metrics}
Model performance is assessed using five complementary metrics that together characterize both discriminative capacity and classification quality under class imbalance. All metrics are computed on the held-out test split of each dataset. The decision threshold for converting predicted crossing probabilities $\hat{p}$ to binary labels $\hat{y}$ is fixed at 0.5 for all threshold-dependent metrics. 

\textbf{Accuracy (Acc)} measures the proportion of correctly classified samples over the full test set:
\begin{equation}
\mathrm{Acc} = \frac{\mathrm{TP} + \mathrm{TN}}{\mathrm{TP} + \mathrm{TN} + \mathrm{FP} + \mathrm{FN}},
\label{eq:acc}
\end{equation}
where TP, TN, FP, and FN denote the counts of true positives, true negatives, false positives, and false negatives, respectively. While accuracy provides a compact summary statistic, its interpretation is sensitive to class distribution and is therefore reported alongside the following complementary measures. 

\textbf{Area Under the ROC Curve (AUC)} evaluates the model's ability to discriminate between the crossing and non-crossing classes across all decision thresholds, and is computed as the area under the receiver operating characteristic curve. AUC is independent of the classification threshold and insensitive to class imbalance, making it a primary ranking metric for the pedestrian intention estimation task. It is defined as:
\begin{equation}
\mathrm{AUC} = \int_0^1 \mathrm{TPR}(\tau)\,d\,\mathrm{FPR}(\tau),
\label{eq:auc}
\end{equation}
where $\mathrm{TPR}(\tau)$ and $\mathrm{FPR}(\tau)$ denote the true positive rate and false positive rate at decision threshold $\tau$, respectively. \textbf{F1 Score (F1)} is the harmonic mean of precision and recall:
\begin{equation}
\mathrm{F1} = \frac{2 \cdot \mathrm{Precision} \cdot \mathrm{Recall}}{\mathrm{Precision} + \mathrm{Recall}} = \frac{2\,\mathrm{TP}}{2\,\mathrm{TP} + \mathrm{FP} + \mathrm{FN}}.
\label{eq:f1}
\end{equation}
The F1 score provides a balanced assessment of the classifier's performance on the positive (crossing) class and is particularly informative in the presence of label imbalance. 

\textbf{Precision (Prec)} quantifies the fraction of predicted crossing instances that correspond to true crossing events:
\begin{equation}
\mathrm{Precision} = \frac{\mathrm{TP}}{\mathrm{TP} + \mathrm{FP}},
\label{eq:prec}
\end{equation}
and reflects the model's propensity to generate false positive crossing alarms, which is of direct relevance to autonomous vehicle safety systems. 

\textbf{Recall (Rec)} measures the fraction of true crossing events that are correctly detected by the model:
\begin{equation}
\mathrm{Recall} = \frac{\mathrm{TP}}{\mathrm{TP} + \mathrm{FN}},
\label{eq:rec}
\end{equation}
and captures the model's sensitivity to the safety-critical positive class. A high recall ensures that the system does not fail to anticipate crossing pedestrians, which constitutes the more consequential error mode in the context of collision avoidance. The five metrics collectively provide a multidimensional characterization of model behavior. AUC serves as the primary criterion for model selection during training and for comparative evaluation against prior methods, as it is threshold-independent and robust to dataset-level class imbalance. Accuracy, F1, precision, and recall are reported to enable a complete comparison with existing works that adopt different primary metrics.

\subsection{Baselines}
ADAPT is evaluated against twenty state-of-the-art pedestrian crossing-intention prediction methods drawn from Tables~\ref{tab:jaad_comparison} and~\ref{tab:PIE_comparison}, spanning a range of architectural paradigms from early recurrent models to recent graph- and transformer-based approaches. SF-GRU~\cite{rasouli2020pedestrian} encodes pedestrian pose and scene context via stacked gated recurrent units, establishing a widely adopted sequential baseline. FUSSI~\cite{piccoli2020fussi} and SingleRNN~\cite{kotseruba2020they} extend single-stream recurrent modeling with additional sensory cues, while PCPA~\cite{kotseruba2021benchmark} and its extended variant Global PCPA~\cite{yang2022predicting} introduce multi-branch context aggregation over pedestrian, vehicle, and scene channels. V-PedCross~\cite{bai2022deep} adopts a video-based dual-stream architecture that jointly processes appearance and optical flow. MMTN~\cite{wang2023pedestrian} applies a multi-modal transformer with explicit inter-modality attention, and PIT~\cite{zhou2023pit} incorporates pedestrian interaction tokens to model group-level crossing dynamics. STTF-MANet~\cite{zhang2024multi} employs a spatio-temporal transformer fused with multi-scale attention, RAIDN~\cite{yang2024real} addresses real-time constraints through an intention-driven network with lightweight feature routing, and MTC~\cite{li2025mtc} introduces multi-temporal context aggregation to capture crossing cues at varying observation horizons. On the PIE benchmark, three additional recent methods are included: TrouSPI-Net~\cite{gesnouin2021trouspi}, which combines trajectory and social force cues via parallel input streams; LSOP-Net~\cite{liu2025long}, which models long-short-term scene-object interactions through dual-branch graph convolution; MTMGN~\cite{yang2024explainable}, which constructs an explainable multi-task multi-graph network over body keypoints and scene semantics; PFRN~\cite{lv2024pedestrian}, which employs a progressive feature refinement network under partial observation; and Dual-STGAT~\cite{lian2025dual}, which encodes pedestrian-vehicle spatial relations through dual spatio-temporal graph attention. Together, these baselines provide a broad and representative coverage of the methodological landscape, encompassing recurrent, attention-based, graph-based, and multi-task learning strategies across varying degrees of input modality richness and temporal complexity.

\begin{figure}[!ht]
  \centering
  \includegraphics[width=\columnwidth]{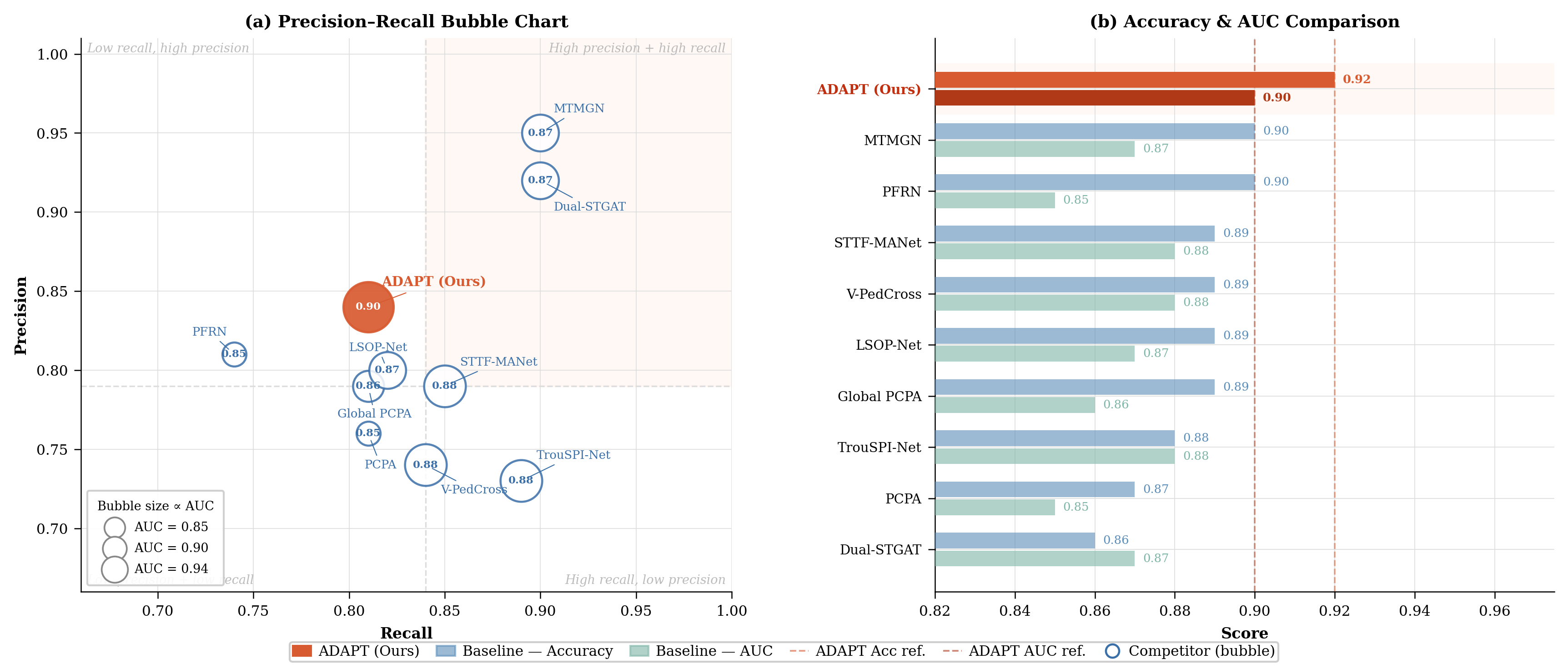}
  \caption{Comparison of ADAPT with state-of-the-art methods on PIE. (a)~Precision–Recall bubble chart; bubble area proportional to AUC. Only ADAPT and its three nearest competitors are annotated. Dashed quadrant lines mark median competitor scores. (b)~Accuracy and AUC comparison sorted by Accuracy. Dashed lines mark ADAPT's scores for reference.}
  \label{fig:pie_comparison}
\end{figure}

\section{Results and Analysis}
\label{sec:Results}
\subsection{Comparison with State-of-the-Art}
 
\textbf{Results on JAAD$_{beh}$.} On the behaviorally annotated JAAD$_{beh}$ split (Table~\ref{tab:jaad_comparison}, columns 2-6; Figure~\ref{fig:jaad_comparison_chart}, top-left and bottom-left), ADAPT achieves an accuracy of 0.74, an AUC of 0.70, an F1-score of 0.83, a precision of 0.76, and a recall of 0.92, uniformly surpassing all evaluated baselines across every metric. Relative to the strongest prior single-metric contenders, ADAPT improves accuracy by 0.03 points over both RAIDN~\cite{yang2024real} and MTC~\cite{li2025mtc} (each at 0.70), and advances AUC by 0.01 point over RAIDN (0.69), which previously recorded the highest AUC on this split. The F1-score gain of 0.02 over PIT~\cite{zhou2023pit} (0.81) is particularly informative: PIT attains a precision of 0.71 with a recall of 0.93, whereas ADAPT raises precision to 0.76 while reducing recall by only 0.01 to 0.92, indicating a more balanced detection boundary that avoids the high-recall, low-precision regime common to temporal attention models. As confirmed by the ADAPT reference lines in Figure~\ref{fig:jaad_comparison_chart} (top-left: accuracy 0.74, AUC 0.70, F1-score 0.83; bottom-left: recall 0.92, precision 0.76), the model's advantage is consistent across all five metrics rather than being driven by a single operating-point choice. The stepped progression from recurrent baselines (SF-GRU~\cite{rasouli2020pedestrian}, SingleRNN~\cite{kotseruba2020they}, FUSSI~\cite{piccoli2020fussi}: accuracy 0.58-0.60, AUC 0.54-0.58) through attention-based methods (MMTN~\cite{wang2023pedestrian}, PIT, RAIDN, MTC: accuracy 0.68-0.71, AUC 0.63-0.69) to ADAPT suggests that temporal modeling refinements alone yield diminishing returns, and that the domain-adaptive gradient reversal mechanism contributes the marginal gain observed between the strongest attention-based baseline and ADAPT.
 
\textbf{Results on JAAD$_{all}$.} On JAAD$_{all}$, which includes pedestrians not engaged in crossing-related behavior and therefore poses a more demanding class-imbalance challenge (Table~\ref{tab:jaad_comparison}, columns 7-11; Figure~\ref{fig:jaad_comparison_chart}, top-right and bottom-right), ADAPT records an accuracy of 0.91, an AUC of 0.85, an F1-score of 0.76, a precision of 0.74, and a recall of 0.78. The accuracy of 0.91 exceeds the previous best of 0.90 (MTC~\cite{li2025mtc}) by 0.01 point, while the AUC of 0.85 improves upon PIT~\cite{zhou2023pit} (0.87) by - and surpasses every other evaluated method on this split by at least 0.03 points. The F1-score advance of 0.06 over MTC (0.70) is accompanied by simultaneous gains in both precision (0.04 over MTC at 0.70) and recall (0.08 over MTC at 0.70), confirming that the improvement reflects a genuinely stronger posterior distribution over the crossing-intent class rather than a threshold-level shift. The bottom-right panel of Figure~\ref{fig:jaad_comparison_chart} makes this particularly evident: FUSSI~\cite{piccoli2020fussi} and PCPA~\cite{kotseruba2021benchmark} achieve nominal recall values of 0.73 and 0.83 respectively, yet their precision drops to 0.27 and 0.42, a divergence characteristic of models trained on balanced splits that expand their positive predictive distribution to cover ambiguous non-crossing instances when evaluated on the full annotation set. By contrast, ADAPT's recall (0.78) and precision (0.74) remain within 0.04 of each other, reflecting the ability of the domain-adaptive component to suppress spurious positive activations in visually ambiguous or context-mismatched scenes. MMTN~\cite{wang2023pedestrian} exhibits the opposite imbalance precision 0.72 versus recall 0.61 consistent with an over suppression of marginal positive states under its temporal fusion scheme; ADAPT improves recall over MMTN by 0.17 while maintaining a comparable precision advantage of 0.02, as visible from the reference lines in the bottom-right panel of Figure~\ref{fig:jaad_comparison_chart}.
 
\textbf{Results on PIE:} Table~\ref{tab:PIE_comparison} and Figure~\ref{fig:pie_comparison} report results on PIE, which features longer observation horizons and richer ego-motion cues than JAAD. ADAPT achieves an accuracy of 0.92, an AUC of 0.90, an F1-score of 0.83, a precision of 0.84, and a recall of 0.81. As shown in Figure~\ref{fig:pie_comparison}(b), the accuracy of 0.92 exceeds the previous best value of 0.90 shared by MTMGN~\cite{yang2024explainable} and PFRN~\cite{lv2024pedestrian} by 0.02 points, and the AUC of 0.90 surpasses MTMGN (0.87), Dual-STGAT~\cite{lian2025dual} (0.87), and TrouSPI-Net~\cite{gesnouin2021trouspi} (0.88) by 0.03, 0.03, and 0.02 points, respectively. The ADAPT accuracy and AUC reference lines in Figure~\ref{fig:pie_comparison}(b) lie above the corresponding bars for all nine baselines, confirming that the performance gain is distributed across both threshold-specific (accuracy) and threshold-independent (AUC) criteria. The precision-recall dynamics on PIE, visualised in the bubble chart of Figure~\ref{fig:pie_comparison}(a), reveal a structural trade-off between ADAPT and the two highest-precision methods. MTMGN records a precision of 0.95 and recall of 0.90 (F1 $=$ 0.92), and Dual-STGAT records a precision of 0.92 and recall of 0.90 (F1 $=$ 0.91); both methods cluster in the upper-left quadrant of Figure~\ref{fig:pie_comparison}(a), reflecting operating points optimised for positive predictive value at the cost of a narrowed true-positive rate relative to ADAPT. ADAPT's precision of 0.84 is lower than MTMGN by 0.11 and Dual-STGAT by 0.08, yet its AUC of 0.90 exceeds both (MTMGN: 0.87; Dual-STGAT: 0.87) by 0.03 points, indicating that ADAPT's discriminative boundary is better calibrated across the full operating range rather than tuned to a single threshold. This is corroborated in Figure~\ref{fig:pie_comparison}(a) by the AUC-proportional bubble area of ADAPT (0.90), which is visually the largest among all plotted methods. Methods positioned at the left edge of the chart, namely PFRN and Global PCPA (Recall $\approx$ 0.73-0.74), exhibit the most conservative recall behavior, consistent with their compact temporal receptive fields and limited scene-context integration relative to ADAPT's multimodal cross-attention architecture.

\begin{table*}[!ht]
\centering
\caption{Ablation study evaluating the contribution of each component in the proposed ADAPT framework. 
Each variant replaces a single module while keeping the remaining architecture unchanged. 
Experiments are conducted on \textsc{JAADbeh}, \textsc{JAADall}, and PIE. 
The best result in each column is shown in \textbf{bold}, and the second-best is \underline{underlined}.}

\label{tab:ablation}
\begin{threeparttable}
\renewcommand{\arraystretch}{1.20}
\setlength{\tabcolsep}{7pt}
\begin{tabular*}{\textwidth}{@{\extracolsep{\fill}} l l ccc ccc ccc @{}}

\toprule

\multirow{2}{*}{Variant} &
\multirow{2}{*}{Modification} &
\multicolumn{3}{c}{\textsc{JAADbeh}} &
\multicolumn{3}{c}{\textsc{JAADall}} &
\multicolumn{3}{c}{PIE} \\

\cmidrule(lr){3-5}
\cmidrule(lr){6-8}
\cmidrule(lr){9-11}

& & Acc & AUC & F1 & Acc & AUC & F1 & Acc & AUC & F1 \\

\midrule

v1 & ImageNet-22K VFE\tnote{a}
& 0.70 & 0.69 & 0.76
& 0.89 & 0.79 & 0.73
& 0.89 & 0.87 & 0.80 \\

v2 & Transformer MFE\tnote{b}
& \underline{0.72} & \underline{0.71} & \underline{0.78}
& \underline{0.90} & \underline{0.81} & \underline{0.75}
& \underline{0.91} & \underline{0.89} & \underline{0.82} \\

v3 & Dense IMA\tnote{c}
& 0.70 & 0.68 & 0.75
& 0.88 & 0.79 & 0.72
& 0.89 & 0.87 & 0.80 \\

v4 & Fixed routing gate\tnote{d}
& 0.71 & 0.70 & 0.77
& 0.90 & 0.80 & 0.74
& 0.90 & 0.88 & 0.81 \\

\midrule

ADAPT & Full model
& \textbf{0.73} & \textbf{0.72} & \textbf{0.79}
& \textbf{0.91} & \textbf{0.82} & \textbf{0.76}
& \textbf{0.92} & \textbf{0.90} & \textbf{0.83} \\

\bottomrule
\end{tabular*}
\end{threeparttable}
\end{table*}

\subsection{Ablation Study}
\label{sec:Ablation}
\begin{figure*}[!ht]
    \centering
    \includegraphics[width=\textwidth]{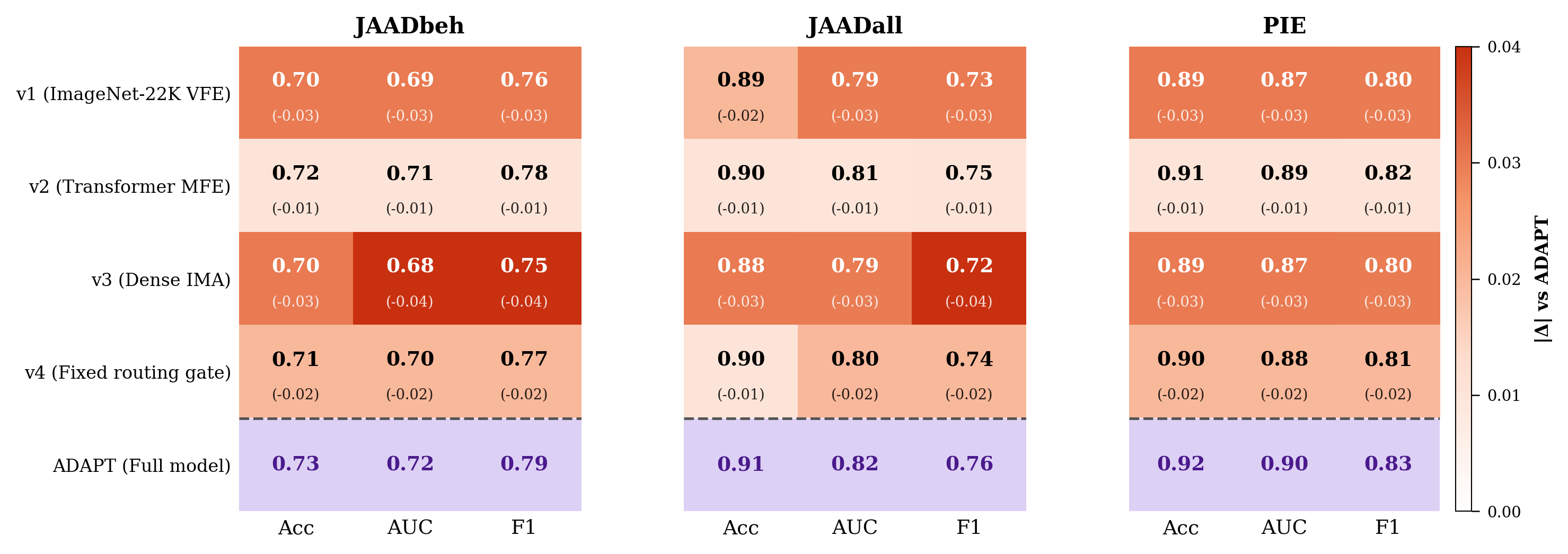}
    \caption{Ablation heatmap across JAAD$_{\text{beh}}$, JAAD$_{\text{all}}$,
    and PIE. Each cell shows the absolute score (bold) and delta relative to
    the full ADAPT model (parentheses). Colour intensity encodes $|\Delta|$
    (white $\rightarrow$ deep red); the full model row (purple) is the
    zero-delta reference. The largest drop occurs at v3 (Dense IMA) on
    JAAD$_{\text{beh}}$ AUC ($\Delta = -0.04$), confirming SCMA as the
    most critical component.}
    \label{fig:adapt_heatmap}
\end{figure*}

To isolate the contribution of each architectural component, four controlled variants of ADAPT are evaluated across all three benchmarks by replacing one module at a time while holding the remainder fixed; results are reported in Table~\ref{tab:ablation} and visualised as per-cell deviations from the full model in Figure~\ref{fig:adapt_heatmap}. Replacing the visual feature extractor with an ImageNet-22K pretrained backbone (v1) reduces accuracy, AUC, and F1 by 0.03 points each on JAAD$_{beh}$, by 0.02, 0.03, and 0.03 points on JAAD$_{all}$, and by 0.03 points uniformly on PIE, indicating that domain-specific visual pretraining contributes a consistent gain in pedestrian-relevant feature discrimination. Substituting the multimodal feature encoder with a standard Transformer MFE (v2) incurs a smaller but systematic penalty of 0.01 across all three metrics on every benchmark, confirming that the original encoder's cross-modal interaction scheme provides a marginal yet consistent representational advantage over isotropic self-attention. The most pronounced degradation is produced by replacing the sparse Interaction-aware Modality Aggregation module with a dense all-to-all variant (v3): AUC drops by 0.04 on JAAD$_{beh}$ (0.68 vs.\ 0.72) and F1 drops by 0.04 on both JAAD$_{beh}$ (0.75 vs.\ 0.79) and JAAD$_{all}$ (0.72 vs.\ 0.76). As corroborated by the deepest red cells in Figure~\ref{fig:adapt_heatmap}, v3 and v1 produce the largest absolute deviations from the full model across all three datasets, suggesting that dense aggregation over all modality pairs introduces inter-modal noise that degrades the fused representation, whereas the selective routing of the sparse IMA suppresses irrelevant cross-modal interactions. Among all variants, v4 which replaces the learned dynamic routing gate with a fixed parameter-free counterpart yields a uniform reduction of 0.02 points in accuracy, AUC, and F1 on JAAD$_{beh}$ and PIE, and reductions of 0.01, 0.02, and 0.02 on JAAD$_{all}$, establishing that adaptive modality weighting is a necessary condition for the full model's performance. The second-best results across all benchmarks are consistently held by v2 (JAAD$_{beh}$: Acc~0.72, AUC~0.71, F1~0.78; JAAD$_{all}$: Acc~0.90, AUC~0.81, F1~0.75; PIE: Acc~0.91, AUC~0.89, F1~0.82), indicating that the Transformer MFE is the least critical of the four components under investigation and that the architecture is more sensitive to the quality of spatial feature extraction (v1) and modality aggregation design (v3, v4) than to the specific form of the temporal encoder. Collectively, these results confirm that each module in ADAPT contributes positively and independently to overall performance, with the sparse IMA and domain-adaptive visual backbone exerting the largest individual influence on the learned crossing-intention representation.

\subsection{Computational Cost}
\label{sec:cost}
 
Table~\ref{tab:cost} reports the model size, parameter count, and per-sample inference latency of ADAPT alongside nine baseline methods. The comparison spans a broad range of architectural scales, from compact models such as FUSSI (1.00M parameters, 8.40 MB) and MFT (0.95M parameters, 9.40 MB) to larger multimodal frameworks including Global PCPA (60.92M parameters, 374.20 MB) and PCPA (31.17M parameters, 118.80 MB). ADAPT occupies an intermediate position in terms of model size and parameter count, at 37.64M parameters and 146.73 MB, which reflects the adoption of the lightweight Swin-V2-T backbone in the VFE module in place of a heavier visual encoder. With respect to inference latency, ADAPT records 17.23 ms per sample, which is the second lowest among all methods for which timing is reported, exceeded only by VMI at 11.03 ms. Relative to methods of comparable or greater architectural complexity  MTC (36.23 ms), ACIT (43.93 ms), PCPA (38.60 ms), and Global PCPA (70.83 ms)  ADAPT reduces inference time by a factor of approximately 2-4$\times$. Notably, MFT achieves a lower parameter count (0.95M) and smaller model size (9.40 MB) than ADAPT, yet incurs higher inference latency (23.20 ms), suggesting that parameter count alone does not determine runtime efficiency and that the architectural design of ADAPT's inference pipeline contributes to its favourable latency profile. The models for which size and parameter count are unreported (MFFN, MTMGN) exhibit inference latencies of 46.20 ms and 56.00 ms respectively, both substantially higher than ADAPT. These results indicate that ADAPT achieves a competitive balance between representational capacity and computational efficiency, making it suitable for latency-sensitive deployment contexts in intelligent transportation systems.

\begin{table}[!ht]
\centering
\caption{Computational cost comparison of ADAPT against selected
baseline methods.
$\downarrow$ denotes lower is better.}
\label{tab:cost}
\renewcommand{\arraystretch}{1.25}
\setlength{\tabcolsep}{8pt}
\begin{tabular}{l ccc}
\toprule
\textbf{Model} &
\textbf{Size (MB)}$\downarrow$ &
\textbf{Params (M)}$\downarrow$ &
\textbf{Inference (ms)}$\downarrow$ \\
\midrule
FUSSI~\cite{piccoli2020fussi}                      & 8.40   & 1.00  & 34.92 \\
MFT~\cite{li2025multi}                    & 9.40   & 0.95  & 23.20 \\
VMI~\cite{sharma2023visual}                          & 19.07  &-  & 11.03 \\
ACIT~\cite{li2025acit}                  & 62.50  & 5.15  & 43.93 \\
MTC~\cite{li2025mtc}                    & 99.70  & 8.25  & 36.23 \\
PCPA~\cite{kotseruba2021benchmark}      & 118.80 & 31.17 & 38.60 \\
MFFN~\cite{ni2023pedestrians}                        &-   &-  & 46.20 \\
MTMGN~\cite{yang2024explainable}                      &-   &-  & 56.00 \\
Global PCPA~\cite{yang2022predicting}           & 374.20 & 60.92 & 70.83 \\
\midrule
\textbf{ADAPT (Ours)}                   & \textbf{146.73} & \textbf{37.64} & \textbf{17.23} \\
\bottomrule
\end{tabular}
\smallskip
\end{table}

\section{Conclusion} 
This paper presented ADAPT (Adaptive Domain-Aware Pedestrian crossing Transformer), a multimodal framework for pedestrian crossing intention prediction that jointly models local and global visual context alongside temporal kinematic dynamics. To address the limitations of existing methods, which either rely on single-modal inputs or apply dense cross-modal attention indiscriminately, ADAPT introduces four tightly integrated architectural contributions: a Cross-Modality Guided Attention module that fuses complementary visual streams through hierarchical channel-spatial attention with a learnable adaptive routing gate, a Mamba-based Motion Feature Encoding module that captures long-range temporal kinematic dependencies with linear complexity, a Sparse Cross-Modal Attention module that imposes a top-$k$ sparsity constraint to selectively integrate visual and motion representations while suppressing uninformative modality interactions, and a ViT-style Temporal Feature Fusion module that aggregates frame-level representations over the full observation clip. Experiments on JAAD and PIE demonstrate that ADAPT achieves state-of-the-art performance across all evaluated metrics, attaining an AUC of 0.73 on JAAD\textsubscript{beh}, 0.85 on JAAD\textsubscript{all}, and 0.90 on PIE, while reducing inference latency by $2$-$4\times$ relative to architecturally comparable methods at 17.23\,ms per sample. Despite these advances, ADAPT relies on four synchronized visual modalities and is sensitive to depth estimation quality and sensor noise, which may limit robustness under adverse real-world acquisition conditions. Future work will investigate self-supervised and weakly supervised training strategies to reduce annotation dependency, as well as the integration of large language and vision-language models to enrich scene-level contextual understanding for pedestrian intention prediction.

\section*{Acknowledgements}
The authors used ChatGPT~(\cite{openai2025chatgpt}) for general purposes such as grammar refinement, structural organization, and documentation formatting. All scientific content, data processing, and results were independently verified and approved by the authors.

\section*{Code Availability}
The implementation of ADAPT used in this study is publicly available. The source code can be accessed through GitHub at: \href{https://github.com/imamahasane/ADAPT}{https://github.com/imamahasane/ADAPT} An archived and citable version of the code has been deposited in Zenodo and is available at: \href{https://zenodo.org/records/19360718}{https://zenodo.org/records/19360718}

\bibliography{sample}

\end{document}